\documentclass[lettersize,journal]{IEEEtran}
\usepackage{amsmath,amsfonts}
\usepackage{algorithm}
\usepackage{algpseudocode}
\usepackage{array}
\usepackage[caption=false,font=normalsize,labelfont=sf,textfont=sf]{subfig}
\usepackage{textcomp}
\usepackage{stfloats}
\usepackage{url}
\usepackage{verbatim}
\usepackage{graphicx}
\usepackage{booktabs}
\usepackage{cite}
\usepackage{hyperref}
\usepackage{multirow}

\usepackage{color, xcolor, soul}

\begin{document}

\title{MistralBSM: Leveraging The Mistral-7B LLM for Vehicular Networks Misbehavior Detection}

\author{\IEEEauthorblockN{\textsuperscript{} Wissal Hamhoum  and Soumaya Cherkaoui} \\
\IEEEauthorblockA{\textit{Department of Computer and Software Engineering} \\
\textit{Polytechnique Montréal, Canada}\\
\{wissal.hamhoum, soumaya.cherkaoui\}@polymtl.ca}
}

\maketitle

\begin{abstract}
Malicious attacks on vehicular networks pose a serious threat to road safety as well as communication reliability. A major source of these threats stems from misbehaving vehicles within the network. To address this challenge, we propose a Large Language Model (LLM)-empowered Misbehavior Detection System (MDS) within an 
edge-cloud detection framework. Specifically, we fine-tune Mistral-7B, a compact and high-performing LLM, to detect misbehavior based on Basic Safety Messages (BSM) sequences as the edge component for real-time detection, while a larger LLM deployed in the cloud validates and reinforces the edge model's detection through a more comprehensive analysis. 
 By updating only 0.012\% of the model parameters, our model, which we named MistralBSM, achieves 98\% accuracy in binary classification and 96\% in multiclass classification on a selected set of attacks from VeReMi dataset, outperforming LLAMA2-7B and RoBERTa. 
Our results validate the potential of LLMs in MDS, showing a significant promise in strengthening vehicular network security to better ensure the safety of road users.

\end{abstract}


\section{Introduction}
\label{sec:introduction}
The evolution of vehicular networks \cite{8300313}, commonly known as vehicle-to-everything (V2X), has been marked by  continuous developments in communication, transportation, and  consumer technologies. Originally conceived as a means of improving vehicle safety through the exchange of information between vehicles and the infrastructure, research expanded to include more advanced scenarios such as cooperative, automated driving and more consumer-centric applications such as in-vehicle infotainment systems, personalized mobility experiences, and seamless connectivity with consumer personal devices \cite{R2023100638}.
This evolution has been accelerated by the integration of emerging technologies like 5G and beyond \cite{9497103} and the Internet of Things (IoT) \cite{9740504}, which enable more sophisticated, consumer-oriented services. 

Today, as the potential of connected autonomous vehicles promises to reshape the landscape of personal and public transportation, the evolution of vehicular networks research continues apace, focusing on enhancing consumer experiences. This includes addressing network security challenges \cite{guan2022overview}, enabling a reliable and secure experience within the intelligent transportation scene.

Indeed, vehicular networks are exposed to various threats resulting from malicious attacks \cite{rizvi2017threat}. These threats compromise the security and reliability of communications among road users, thereby jeopardizing road and traffic safety. 
One of the main vectors of these attacks within vehicular networks is vehicle misbehavior \cite{10201849}. Misbehaving vehicles can launch denial-of-service (DoS) attacks, which flood the network with malicious traffic, disrupting communication between vehicles \cite{fotohi2020new}. As a result, this disruption can lead to communication breakdowns, making vehicles unaware of each other and disrupting the operation of both traffic systems and vehicular safety applications. In addition, message tampering and integrity attacks can mislead on-board control systems, compromising vehicle safety. In this scenario, misbehaving vehicles deliberately transmit inaccurate or erroneous messages despite normal hardware and software functioning \cite{10026338}. Vehicle misbehavior can result in significant repercussions, including serious traffic disruptions or even accidents. Hence, detecting misbehavior is crucial in vehicular communication systems. However, it is a complex challenge due to the dynamic nature of traffic environments and the diverse forms misbehavior can take. Vehicles interact unpredictably, leading to various behaviors that may differ from expected patterns. Additionally, the large volume of data generated by vehicles increases the complexity. Analyzing this data in real-time requires advanced algorithms and computing resources.

Machine learning (ML) has recently emerged as an interesting solution in the development of Misbehavior Detection Systems (MDS). ML enables MDS to adapt and learn from historical data, continuously improving its ability to identify abnormal or malicious behavior \cite{boualouache2023survey}. 
With the recent advancements in large language models (LLMs) there is an exciting opportunity to further bolster the capabilities of MDS. Indeed, models such as the GPTs \cite{yenduri2023generative}, LLAMA-2\cite{touvron2023llama} and BERT \cite{devlin2019bert} have demonstrated remarkable capabilities in various natural language processing tasks  but also showed interesting capacities in network security tasks \cite{chen2022bert, 10414427,9534113 }, including for vehicular networks' security \cite{10272465, 10069190}.

The potential of LLMs in vehicle misbehavior detection stems from the fact that LLMs can facilitate the extraction of insights from textual data of vehicle communications. Through semantic analysis and contextual understanding, these models can identify relevent features that can signify potential abnormal behavior patterns. However, using LLMs for vehicle misbehavior detection poses significant challenges, notably in privacy and resource constraints. Given their extensive parameter counts, often in the tens of billions or even trillions, LLMs are naturally suited for cloud deployment. Nevertheless, transferring all vehicle communication data to cloud infrastructure poses potential risks for real-time misbehavior detection and raises concerns about data privacy.

 Recently, several LLMs were introduced, among which  Mistral-7B \cite{jiang2023mistral} has gained attention for its exceptional balance between model size and performance, making it particularly suitable for ressource constrainted application like vehicular networks. More importantly, it is open-weight model, which opens the possibility to adapt and leverage it as an LLM-empowered MDS architecture. In this work we propose an edge-cloud setting, where a quantized  Mistral-7B is deployed at the edge (e.g., at a roadside unit (RSU), and a deeper LLM resides in the cloud. The Edge LLM can perform initial vehicle misbehavior detection in real-time on short Basic Safety Message (BSM) sequences, providing immediate feedback and alerts to vehicle systems. Cloud LLMs can complement this by conducting more in-depth analysis on longer BSM sequences and by incorporating additional contextual information from a broader and possibly multi-modal dataset stored in the cloud (e.g., multimodal traffic information from the intelligent transportation system). Using the edge LLM has the advantage of processing vehicular communication data locally, minimizing the need to transmit raw data to the cloud and preserving user privacy. Cloud LLMs can be utilized for more general analysis that uses edge LLMs insights and other data external to the communication system, ensuring compliance with privacy regulations and minimizing the risk of data privacy breaches.

In this article, we focus mainly on the implementation of Mistral-7B as the edge LLM to classify vehicle behaviors. Thus, the main contributions of this article are as follows: 
\begin{itemize} 
    \item We propose an LLM-based Misbehavior detection framework leveraging Mistral-7B for BSM sequence classification. 
    \item We design our data preprocessing carefully to avoid label leaking or training data leaking for a more reliable performance assessment and efficient model inference to enable easy edge deployment.
    \item We compare Mistral-7B's performance to other prominent open-source LLMs, namely LLAMA2-7B and BERT-based architecture, in the context of MDS, and we demonstrate its superior results.
    \item We use quantization and study BSM sequence lengths to optimize model deployment within the limited computational resources of vehicular networks.
\end{itemize}

The rest of this paper is organized as follows. Section \ref{sec:relatedWorks} introduces the related works of misbehavior detection methods in vehicular networks. Section \ref{sec:background} introduces a background about vehicular networks' architecture, LLMs and low-rank adaptation for LLM fine-tuning. In Section \ref{sec:systemModel}, we define our system model  and introduce Mistral for misbehavior detection. In Section \ref{sec:EvalPer}, we do a performance evaluation of our model and compare our results.

\section{Related works}
\label{sec:relatedWorks}
Misbehavior detection is a pivotal component in vehicular networks where ML can offer an advantage over rule-based approaches given the significant amount of communication data generated by the network and required to achieve an accurate detection. In the literature, several ML-approaches have been proposed, from simple algorithms to deep models, to address in-vehicle attacks as well as V2X communications misbehavior. For instance, the authors \cite{10307254} studied  several famous ML algorithms, including tree-based models such as Decision Tree, XGBoost, and Ada Boost classifiers in addition to Ridge classifier and Naive Bayes on CAN-OITDS dataset, where XGBoost achieved the best performance. On the other side, the authors of \cite{9245568} studied ML for V2X anomaly detection using models such as Support Vector Machines (SVM), K-Nearest Neighbors (KNN), Naive Bayes, Random Forest, ensemble boosting, and voting ensemble, and were able to achieve good classification results. Despite being lightweight and achieving promising results, simple ML-techniques lack the ability to process the time facet of network data. 

This drawback was addressed by LSTM-based architectures that were designed to effectively model temporal data. In ~\cite{9641946, 10.1145/3491396.3506509,alladi2023deep }, the authors combined LSTM and CNN to derive spatial and temporal features, achieving good misbehavior detection. Additionally, the time aspect was further emphasized in \cite{10001264} where the authors approached misbehavior detection as a time-series anomaly detection problem, modeled as a Markov Decision Process (MDP), where each decision depends on preceding states. The improvement introduced by these approaches highlights the importance of the temporal dimension in detecting misbehavior within vehicular networks. 

The emergence of transformers shifted the interest due to their unprecedented capabilities to efficiently process sequenced data types. In vehicular networks, CAN-bus anomaly detection took advantage of these capabilities of transformer-based models and showed that BERT-inspired architectures surpass other deep models \cite{10069190, alkhatib2022canbert, li2023securebert}.Beyond works focusing exclusively on CAN-bus, the authors in \cite{10533857} propose a hybrid Network IDS (NIDS), named IoV-BERT-IDS, for both in-vehicle (CAN-bus) and extra-vehicle networks. The model leverages a semantic extractor to convert raw CAN-Bus and network sequences into BERT-compatible representations, pretrained using Masked Byte Word Model (MBWM and Next Byte Sentence Prediction (NBSP) tasks to capture contextual and sequential features. It is then fine-tuned for intrusion detection, achieving about 99\% accuracy across CICIDS, BOT-IOT, Car-Hacking, and IVN-IDS datasets. However, these datasets are not specifically V2X communication datasets; as a consequence, we cannot assume systematically that the model will give the same performance in our study case.

On the other hand, the authors in \cite{10272465} focused on V2X communications misbehavior by fine-tuning an encoder-based vanilla transformer in a semi-supervised approach to detect anomalies. Their model reconstructs vehicle behavior by regenerating position and speed from sequences of 200 BSMs, then detects misbehavior based on the Mean Squared Error (MSE) between the original and reconstructed features compared to a threshold, yet their data preparation includes a hint to the vehicle's ID in the dataset, which can lead to data label leakage. Additionally, Larger LLM models such as GPT-4o, LLaMA-3.2 and Gemini 1.5 Pro were used in \cite{10933799} alongside DNNs such as VGG19 and ResNeXt to detect motion forging and natural denoising diffusion (NDD) attacks. The results indicate that GPT-4o delivers the strongest performance overall, though the model is not open source, which in a practical scenario will require sharing sensitive vehicular network telemetry with a third party, raising serious privacy and security concerns. Furthermore, LLMs were coupled with explainable AI technics explainable AI technics such Kappa coefficient, Matthews Correlation Coefficient, and  SHapley Additive exPlanation (SHAP), in \cite{khan2025novel},  to validate the model's misbehavior detection capabilities. This approach studies aims to understand the reason behind a vehicle behavior classification achieved by LLM, however the SHAP results strongly suggest that the message ID feature has the biggest influence on the classification, raising potential overfitting or train/test data leakage. 

From another perspective, the authors in \cite{10283969} propose a transformer-based Intrusion Detection System (IDS) for V2X trained within federated learning approach. Their model is a tailored encoder-based transformer that integrates feature attention and routing to select the most suitable feature set for network packet classification. The proposed system overlooks temporal context of the packet in classification, which can be considered a major drawback. This limitation is addressed in \cite{10921812}, where the authors train their transformer encoder-based model to classify BSM-windows, showing that the federated learning gives a model with better classification accuracy compared to the standalone model. 

 Despite the growing use of LLMs for misbehavior detection, many studies provide limited details on data preparation. The results of some works strongly suggest potential training data leakage, particularly when identifiers such as message ID or pseudo-ID features are used, making it difficult to attribute performance gains solely to the model’s learning capacity. Moreover, it remains crucial to investigate the computational resources required effectively to deploy such models in real-world settings.

\section{Background}
\label{sec:background}
\subsection{Vehicular Communication Models}

\begin{figure*}[!ht]
    \centering
    \includegraphics[width=0.85\linewidth]{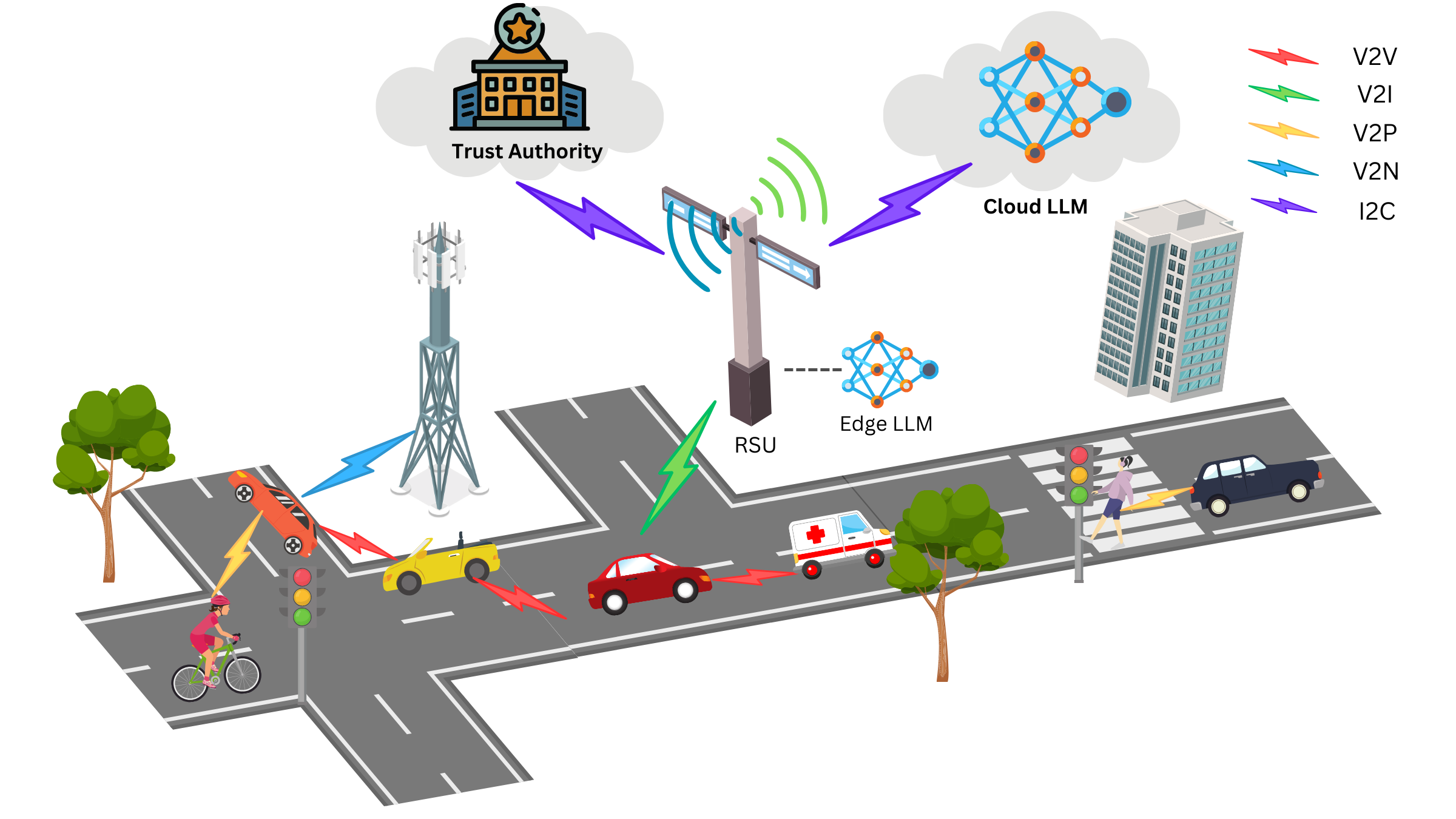}
    \captionsetup{justification=centering}
    \caption{V2X communication and Misbehavior detection with edge and cloud LLMs }\label{fig:typesV2X}
\end{figure*}

V2X communications encompass various technologies facilitating real-time information exchange between vehicles and their surroundings, thereby enhancing road safety, and traffic efficiency. Fig.\ref{fig:typesV2X} provides an overview of the different types.

V2V communication involves vehicles equipped with network modules, called  onboard units, enabling them to exchange data concerning their position, speed, heading, and acceleration. This type of communication aims to prevent accidents by facilitating situational awareness among vehicles.

In contrast, V2I communication allows vehicles to interact with RSUs such as streetlamps, RFID readers, lane markers, and signage. This bidirectional communication enables the real-time exchange of information about road conditions, traffic, accidents, and more, thereby enhancing overall traffic management and safety\cite{fi11020027}.

V2P, or vehicle-to-pedestrian communication, involves interactions with Vulnerable Road Users (VRU), such as pedestrians, cyclists and motor wheelers \cite{9963973}. These individuals typically carry smartphones or wearable devices capable of transmitting their location to alert nearby vehicles of their presence, contributing to enhanced safety for all road users.

Additionally, V2X includes V2N communication, which enables vehicles to connect to cloud-based services and platforms. This connectivity facilitates access to a wide range of services, such as over-the-air updates, remote diagnostics, real-time traffic and navigation information, and data analytics for traffic management purposes.

In order to protect ensure vehicle privacy, vehicles use pseudonyms for communications. Vehicle pseudonyms are temporary, randomly generated identifiers used in vehicular communication systems to provide privacy and anonymity for vehicles while still allowing them to communicate with each other and infrastructure. These pseudonyms are periodically changed to prevent tracking of vehicles over time \cite{8345196}.

\subsection{Transformers and LLMs}
The transformer is a groundbreaking architecture that revolutionized the task of sequenced data processing, introducing innovative mechanisms for capturing complex dependencies and patterns within sequences. This was achieved by leveraging the self-attention mechanism to weight the importance  of each token in sequence compared to the other tokens. The weights are obtained by applying three linear transformations, namely (keys (k), queries (Q) and values (V)) on the input sequence of size $d_k$ following the equation \ref{eq: self-attention}.

\begin{equation}
    \text{Attention (Q, K,V)} = softmax(\frac{QK^T}{\sqrt{d_k}})V 
    \label{eq: self-attention}
\end{equation}

\begin{figure}[!ht]
    \centering
    \includegraphics[width=0.7\linewidth]{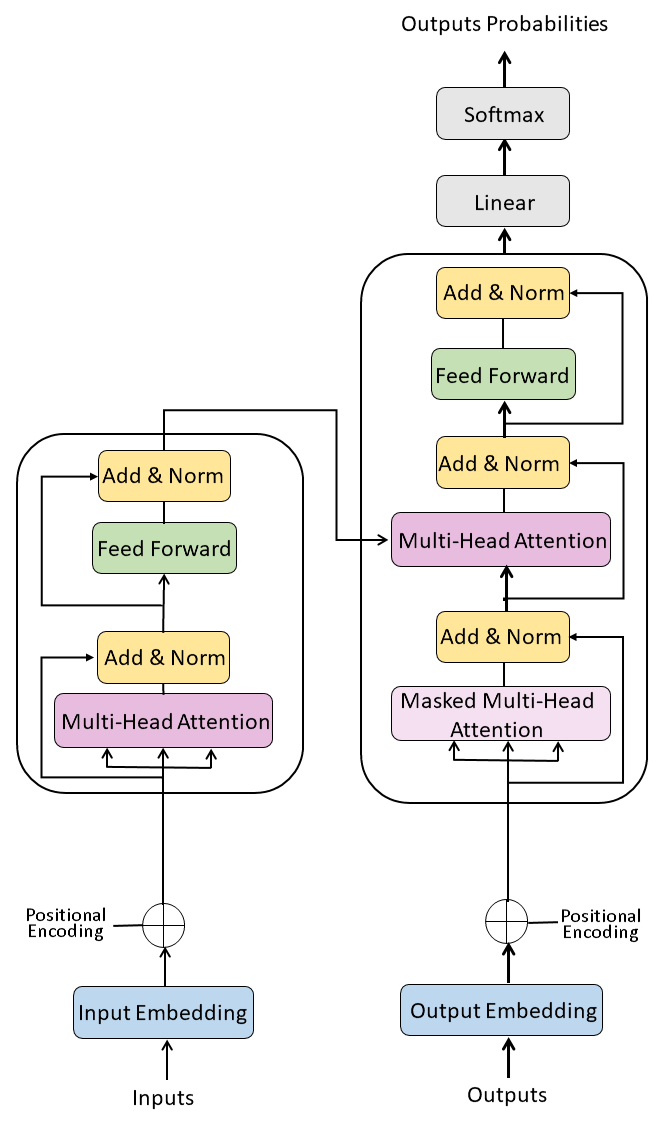}
    \captionsetup{justification=centering}
    \caption{Transformer architecture}\label{fig:LLM}
\end{figure}

Several variations of transformers have been proposed and customized to address a broad spectrum of tasks, encompassing language understanding, generation, translation, and more. Notable examples include the original Transformer architecture \cite{vaswani2023attention} which is presented in Fig. \ref{fig:LLM}, formed by both encoder and decoder blocks. BERT \cite{devlin2019bert} and RoBERTa \cite{liu2019roberta}, are another type of architecture formed only by encoder blocks. On the other hand, LLAMA \cite{touvron2023llama} and GPT are decoder-only types of transformers, primarily designed for generation tasks.
The difference between the encoder-based and decoder-based transformer is self-attention module type. While in the case of BERT-like models, the attention is bidirectional, meaning that all the sequence tokens are used to compute the attention score for each token. The decoder-based models use masked self-attention technique, in which computing the attention score of a token requires only past tokens

Mistral-7B is another example of a decoder-based transformer that leverages specific design decisions, allowing it to outperform models such as the 13 billion-parameter variation of Llama-2 and the 34 billion-parameter Llama-1 with only 7 billion parameters. Those technical choices involve using Sliding Window Attention (SWA) that operates by defining a fixed-size window surrounding each token, thus limiting the number of tokens it has to tend to \cite{beltagy2020longformer}. Compared to the vanilla self-attention used in \cite{vaswani2023attention}, this technique enables longer sequence processing with lower computation cost. Moreover, it employs the Grouped-query attention (GQA) technique, where a group of queries shares a single key and value head \cite{ainslie2023gqa}, thus speeding up the model inference time.

 Mistral-7B is suited for misbehavior detection in vehicular networks because it combines the performance of larger models with a more compact size. Its smaller size and efficient sequence processing make it more suitable for deployment at the edge, where computational resources are often limited \cite{technologies12060081}. Additionally, Mistral-7B’s pretraining on a large and diverse dataset enables easier fine-tuning for specific tasks like ours, requiring less data for adaptation. This advantage reduces the overall cost of supervised learning while leveraging its powerful ability to extract semantic information from sequences. Overall, Mistral-7B’s capabilities make it suitable  for misbehavior detection in vehicular networks. 

\subsection{Low-rank Adaptation}
In traditional fine-tuning, the model's weights are directly modified to adapt the model to the new considered task. While fine-tuning is a crucial approach in ensuring transfer of knowledge, the process becomes computationally daunting as the number of parameters increases. Notably, fine-tuning large language models poses significant challenges in terms of memory requirements and computational resources. 

Low-Rank Adaptation (LoRA) is a technique that was proposed in \cite{hu2021lora} to alleviate the memory footprint and accelerate the process of large models' fine-tuning. This is achieved through reducing the number of trainable parameters by decomposing the changes on the pre-trained weight matrices into low rank matrices. The process supposes that after fine-tuning, the new weight matrices will have the form $W =W + \Delta W$ where W is frozen during training and $\Delta W$ is decomposed into two lower-rank matrices, A and B. Being of lower rank, the number of trainable parameters in these matrices is smaller than the original$\Delta W$. 

Although LoRA enhances significantly the computational requirements, further optimization could be achieved by using QloRA \cite{dettmers2023qlora}, which applies quantization techniques to further reduce the memory footprint.

\section{System model}
\label{sec:systemModel}
We present an approach centered around the utilization of Mistral-7B for detecting vehicle misbehavior. Our method involves reframing the misbehavior detection problem as a sequence classification task. In this framework, we deploy a model based on Mistral-7B, which is quantized and implemented on a RSU. Through quantization, the memory requirements for running Mistral-7B are significantly reduced, enabling its deployment on edge equipment at the RSUs. These RSUs possess the necessary capabilities to collect BSMs exchanged within their sensing ranges and execute the model to identify potential misbehavior.

\begin{figure*}[!ht]
    \centering
    \includegraphics[width=1\linewidth]{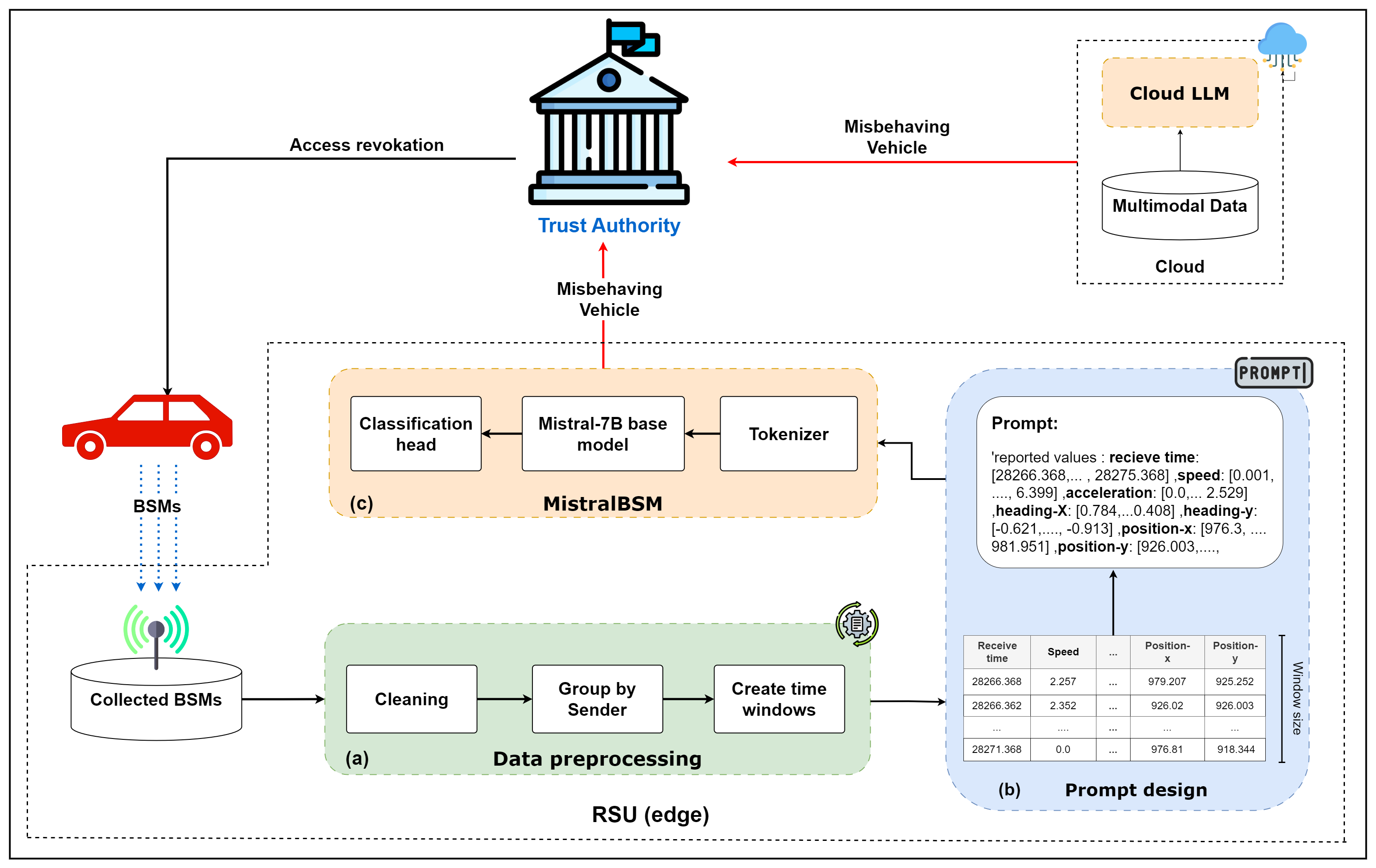}
    \captionsetup{justification=centering}
    \caption{Proposed architecture}
    \label{fig:Prop_architecture}
\end{figure*}

The process of classifying vehicle behaviors consists of three steps, as shown in Fig.\ref{fig:Prop_architecture}. The first step involves data preprocessing, detailed in submodule (a) of Fig.\ref{fig:Prop_architecture}. In this stage, features are extracted from detected BSMs in the RSU's local database, following the data preparation steps detailed in \ref{subsection: Veremi}. Thereafter, the BSM messages are grouped by sender pseudonym and transformed into BSM-Windows of size $n$ for each vehicle. The second step, represented by the prompt design module in Fig.\ref{fig:Prop_architecture}, involves converting multivariate time windows into a textual format. The text structure is as follows: 
\begin{quote}
\texttt{\textbf{reported values}: $Feature_1$:$[value\_1, value\_2, ..., value\_n]$, $Feature2$: $[value\_1, value\_2, ..., value\_n]$, ..., $Feature_M$: $[value\_1, value\_2, ..., value\_n]$}
\end{quote} 

The third and final step is classification, handled by the MistralBSM model in submodule (c) of Fig.\ref{fig:Prop_architecture}. This module consists of a Tokenizer, Mistral-7B layers for prompt processing, and a classification head that assigns a behavior label. Following the classification, the RSU reports the information about misbehaving vehicles is communicated to a Trust Authority (TA) located in the cloud.
 
The TA is, thus, responsible for verifying the reported misbehavior. If further analysis is needed, the TA forwards the report to a cloud-based LLM, which enhances the report by conducting deeper analysis and incorporating additional data from a broader dataset stored in the cloud.
 For instance, the cloud LLM can access historical data on the vehicle’s behavior, as well as sensor data from roadside infrastructure such as LIDAR or radar, allowing for a more thorough assessment of whether the vehicle was transmitting accurate information. This deeper analysis is implemented at the cloud level rather than at the edge for privacy reasons, since only the TA has access to the vehicle's real identity. By performing this task in the cloud, sensitive information about the driver or vehicle remains protected and is only accessed when necessary, such as in cases where misbehavior needs to be verified before revoking network access.
Subsequently, the TA takes the appropriate response that could include revoking the vehicle's access to the network.
In this paper, we assume the existence of such a cloud-based model and focus our study on the edge-LLM operating at the RSU. In fact, MistralBSM's methodology  ensures the examination of both contextual and content aspects within the messages at the edge LLM. Contextual dimensions are considered during the grouping of messages into time windows, facilitating the analysis of temporal patterns and relationships. Simultaneously, the content aspect is considered by a detailed inspection of the message contents.

\section{Experiments and Results}
\label{sec:EvalPer} 
In this section, we delve into the details of the conducted experiments. We start by presenting the preprocessing steps we applied on the database used, namely the VeReMi dataset. Thereafter, we define the performance metrics used to evaluate  our model. Thereafter, we present the two major sets of experiments: binary and multiclass classification. For each set, we investigate two possible methods of leveraging Mistral-7B: one involving fine-tuning and the other without. We aim to investigate the possibility of leveraging Mistral-7B with minimal modifications, maximizing its utility without extensive adjustments. From there, we compare its performance against two other LLMs, LLAMA2-7B and RoBERTa.
Lastly, we analyze the effect of varying window sizes on the inference time to identify the requirements and potential solutions for deploying the models.

\subsection{Extended VeReMi Dataset } 
\label{subsection: Veremi}
VeReMi \cite{van2018veremi} is a  Vehicular Reference of Misbehavior dataset that was created to evaluate the performance of misbehavior detection systems in the VANETs. It was generated using the Framework For Misbehavior Detection (F2MD), which is an extension of the vehicular network simulator VEINS, and a subsection of the Luxembourg SUMO Traffic (LuST) scenario covering 1.61 km² with a peak density of 67.4 vehicles per km². To experiment and assess our approach, we used the extended version of VeReMi\cite{9149132} that offers more diverse attack types, including DoS, Random Positioning Attacks, Constant Positioning Attacks, Disruptive Attacks, Constant Speed Attacks, Delayed Messages Attacks, Data Replay Attacks, Traffic Sybil Attacks, and Eventual Stop Attacks that were simulated under varying traffic conditions, specifically from 07 to 09 am and from 2 to 4 pm. The dataset is structured into trace files containing the exchange of BSMs by individual vehicles. 
During each attack simulation, vehicle behaviors are stored in separate files encompassing various fields including: type, receive time, sending time, sender ID, sender Pseudo, message ID, position, position noise, speed, speed noise, acceleration, acceleration noise, heading, and heading noise.

Preprocessing of the raw files involved extracting all messages, concatenating them, and selecting only type-3 messages, which describe the received BSMs. Duplicate entries resulting from the interception of the same message by multiple vehicles were eliminated. Then fields describing noise attributes such as position noise, heading noise, acceleration noise, and speed noise were dropped.

In order to optimize memory usage during fine-tuning and inference, we made adjustments to some fields. For instance, we calculated the Euclidean norm for speed and acceleration fields in order to compress the 3-value features into a single value. Additionally, the position and heading fields were split into two distinct fields: position-x and position-y, and heading-x and heading-y, respectively. Given that the z coordinate remained consistently null across all entries, it was removed. Lastly, to further reduce memory requirements, all numerical values were rounded to three decimal points. Thereafter, each message is labeled based on its sender ID; therefore, every message issued by a misbehaving car was labeled as attack. This classification is derived from the title of the corresponding trace file, referencing the attack-code correspondence outlined in Table \ref{tab:name_label_mapping}. Once the dataset was labeled, messages for each sender ID were grouped, ordered by the receive time, and segmented into time windows. Importantly, we removed the sender ID, sender Pseudo, message ID features to avoid label leaking during training. 

The final preprocessing step is the textualization of the time windows, which could be achieved either by considering each row as a separate sentence having the  feature values  preceded by the corresponding field name or column-wise, like described in the submodule (b) in Fig. \ref{fig:Prop_architecture}. As the former results in large prompt size due to repeated field names, we resorted to the column-wise approach. 

For our experiments, we consider the two prominent types of misbehaviors in V2X communications: faulty-hardware-like behavior and purely malicious attacks. Therefore, we select Constant Position, Constant Speed, and Eventual Stop as the former type and Data Replay, Dos, Dos Random, and Dos Random Sybil as the malicious attacks from the extended VeRiMi dataset, and we preprocess them to extract BSM sequences as time windows.

\begin{table*}[!ht]
\center 
\caption{Mapping attack type to codes}
\begin{tabular}{|c|l|l|}
\hline
\textbf{Identifier} & \textbf{Attack Type} & \textbf{Attack description}\\ \hline
A0 & Genuine & Normal behaving vehicle with no malicious intention \\ \hline
A1 & ConstPos & Vehicle transmitting constant position attributes  \\ \hline
A5 & ConstSpeed & Vehicle transmitting constant speed field. \\ \hline
A9 & EventualStop & \begin{tabular}[c]{@{}l@{}}Vehicle simulating a sadden stop by freezing its position and setting \\ its speed to zero.\end{tabular} \\ \hline
A11 & DataReply & Vehicle retransmitting messages received from a target vehicle. \\ \hline
A12 & DelayedMessages & Vehicle transmitting delayed correct information. \\ \hline
A13 & DoS & Vehicle transmitting at rate exceeding the predefined limit. \\ \hline
A14 & DoSRandom & Malicious V2X node, conducting DoS with transmitted fields set randomly. \\ \hline
A18 & DoSRandomSybil & Malicious node would use Sybil attack to conduct DoS random.\\ \hline
\end{tabular}
\center
\label{tab:name_label_mapping}
\end{table*}

\begin{figure*}[!ht]
    \centering
     \includegraphics [width=0.8\linewidth]{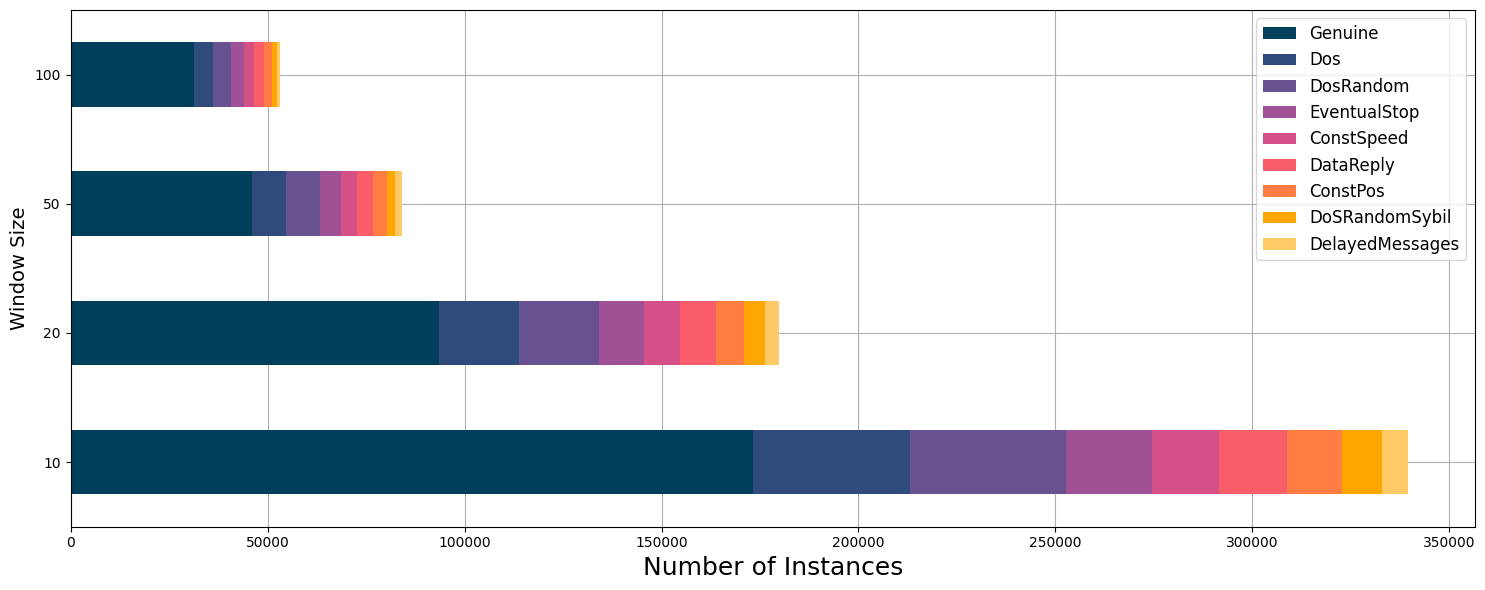}
    \captionsetup{justification=centering}
    \caption{The class distribution after dataset preprocessing}
    \label{fig:data_dist}
\end{figure*}

In order to be able to select what is the best number of packets inside a time window, we created different datasets with 4 sizes, 10, 20, 50 and 100. Table \ref{tab:name_label_mapping} and Fig.\ref{fig:data_dist} give an overview of the selected attacks and class distribution after processing.

\subsection{Performance Metrics}
\noindent
To assess the performance of our misbehavior detector, we used the following metric suit. 
\paragraph{Recall} measures the proportion of actual positives that are correctly identified by the system. A high recall reflects the detector's ability to identify the actual attack and minimize the chances of labeling an attack as normal.
\begin{equation}
    \text{Recall} = \frac{\text{True Positives}}{\text{True Positives} + \text{False Negatives}}
    \label{eq:recall}
\end{equation}
\paragraph{Precision} Calculates the correctly labeled attacks to the actual attacks, forming an idea on the model's capacity to identify attacks while minimizing false alarms.
\begin{equation}
    \text{Precision} = \frac{\text{True Positives}}{\text{True Positives} + \text{False Positives}}
    \label{eq:precision}
\end{equation}

\paragraph{F1 score} It is the harmonic mean average of the recall and precision, allowing to assess the model's capacity to balance the two, thus providing a balanced evaluation of the system's performance.
\begin{equation}
    F1 = 2 \times \frac{\text{Precision} \times \text{Recall}}{\text{Precision} + \text{Recall}}
\end{equation}

\paragraph{Accuracy} Computes the percentage of instances correctly classified by the model. It reflects the model's ability to correctly identify the normal from the anomalous behavior.
\begin{equation}
    \text{Accuracy} = \frac{\text{True Positives} + \text{True Negatives}}{\text{Total Instances}}
\end{equation}

\subsection{Misbehavior Detection}
In the first set of experiments, we utilize Mistal-7B to distinguish between normal behaving cars and misbehaving ones. In the following subsections, we outline the different settings we experimented with Mistral-7B, and we compare its performance against two other renown LLMs. 
\subsubsection{Fine-tuned Mistral}
\begin{figure}[!ht]
    \centering
    \includegraphics[width=0.85\linewidth]{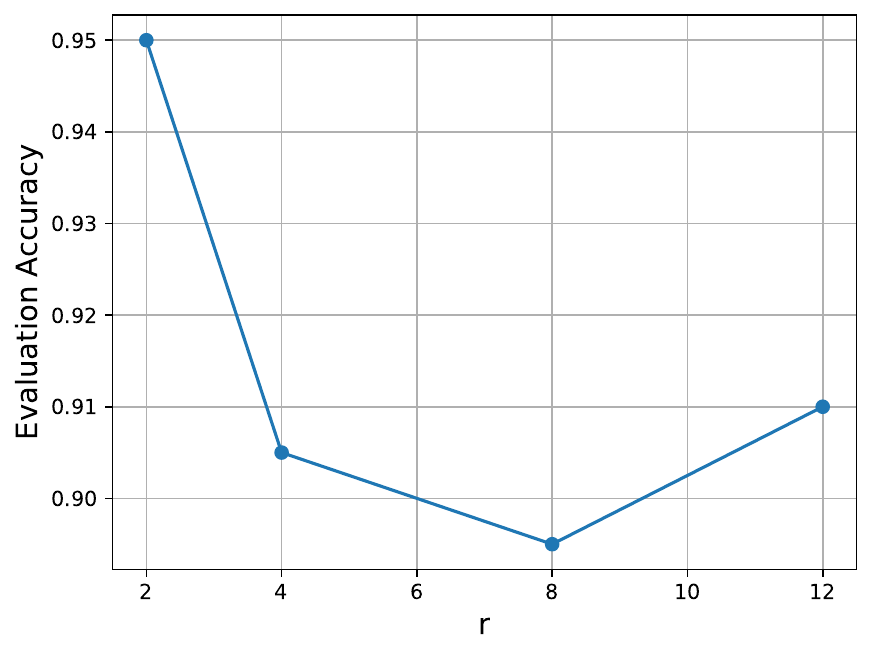}
    \captionsetup{justification=centering}
    \caption{The effect of low-rank matrix rank on the evaluation accuracy.}
    \label{fig:r_rank}
\end{figure}

We employed qLoRA for fine-tuning Mistral-7B, leveraging the hyperparameters outlined in Table  \ref{tab:hyperparameters}. The classifier is formed by Mistral-7B transformer, on top of which a linear classification head based on the last token is added. \\
We chose to perform the fine-tuning on the self-attention modules and the weights of the classification head to maximize performance while minimizing parameter count, as recommended in \cite{hu2021lora}. Furthermore, the $r$ and the LoRA alpha in the Table \ref{tab:hyperparameters} refer to the dimension of the low-rank matrices and the scaling factor which calibrates the weights update magnitude during the training, respectively. Fig. \ref{fig:r_rank} presents the effect of the hyperparameter r on the accuracy when window size equals 10. The figure shows that the best accuracy is obtained for an r value of to 2, whereas it decreases significantly for higher values. This justifies the choice indicated in Table \ref{tab:hyperparameters}. 

The findings presented in Table \ref{tab:Mistral_Binary_results} summarize the outcomes of fine-tuning Mistral-7B. We can notice that, the more the window size increases, the better the model is able to classify vehicle behavior. This is because larger windows offer more information on the messages communicated by the vehicle, making the classification task easier. The most favorable results were achieved using a window size of 100, yielding an accuracy and F1 scores of 98\%. Notably, this was achieved despite the utilization of only 1000 instances per class and tuning only 0.012\% of the parameters.

\begin{table}[!ht]
\centering
\caption{Hyperparameter settings for fine-tuning with QLoRA}
\begin{tabular}{|l|c|}
\hline
\textbf{Hyperparameter} & \textbf{Values} \\ \hline
Quantization type & 4-bit \\ \hline
$r$ & 2 \\ \hline
LoRA alpha & 16 \\ \hline
LoRA dropout & 0.05 \\ \hline
Bias & none \\ \hline
Target modules & \multicolumn{1}{l|}{{[}q\_proj, v\_proj, score{]}} \\ \hline
Task type & sequence classification \\ \hline
\end{tabular}
\label{tab:hyperparameters}
\end{table}

\begin{table}[!ht]
\centering
\caption{Binary Mistral-7B results}
\begin{tabular}{|c|c|c|c|c|}
\hline
Window size & Accuracy & F1 & Recall & Precision \\ \hline
10 & 0.945 & 0.944933 & 0.945445 & 0.948269 \\ \hline
20 & 0.970 & 0.969988 & 0.970297 & 0.971429 \\ \hline
50 & 0.980 & 0.979998 & 0.980198 & 0.980583 \\ \hline
100 & 0.980 & 0.98020 & 0.980120 & 0.98028 \\ \hline
\end{tabular}
\label{tab:Mistral_Binary_results}
\end{table}

\subsubsection{Mistral-7B as Feature Extractor} For this set of experiments, we train a classification neuron network on top of frozen Mistral-7B layers, that we call MistralFeatureNet. The purpose of MistralFeatureNet is to study the capacity of the transformer to extract relevant features from the textualized windows without any fine-tuning.
 The classification process starts by collecting Mistral-7B's top layer output of shape $(n_{tokens}, 4096)$,  then max-aggregating it into a 4096-sized vector. Thereafter, the resulting vector is passed through several dense layers  for the final classification. The number of hidden layers and their specifications are indicated in Table \ref{tab:hyperparmeters_b_classifier}.\\
We performed the training on a balanced basis between normal and abnormal cases. Since size 100 contains the fewest data points, we were careful to select a similar class distribution for the other sizes, i.e. 21923 per class. Table \ref{tab:FMistral_Binary_results} summarizes the results. The table shows the model's positive outcomes, especially  with 10 and 20 window sizes, with an accuracy and F1 above 90\%. 
In addition, we observe that for bigger window sizes, the performance deteriorates. A possible cause of this is the max aggregation step. Since we are taking the maximum over the tokens' representation on each model dimension, the larger the sequence, the higher the chance of loosing information.\\

\begin{table}[!ht]
\centering
\caption{Binary classification head hyperparameters}
\begin{tabular}{|l|l|}
\hline
\textbf{Hyperparameter} & \textbf{Value} \\ \hline
Input dimension & 4096 \\ \hline
Hidden Layers dimension & {[}256, 128, 64, 32{]} \\ \hline
dropout rate & 0.2 \\ \hline
Optimizer & Adam \\ \hline
Learning rate & 0.0001 \\ \hline
Number of epochs & 80 \\ \hline
Batch size & 32 \\ \hline
\end{tabular}
\label{tab:hyperparmeters_b_classifier}
\end{table}

\begin{table}[!ht]
\centering
\caption{Binary MistralFeatureNet results}
\begin{tabular}{|c|c|c|c|c|}
\hline
Window size & Accuracy & F1 & Recall & Precision \\ \hline
10 & 0.94100 & 0.940000 & 0.902000 & 0.980000 \\ \hline
20 & \textbf{0.94200} & \textbf{0.943000} & \textbf{0.930000} & 0.980000 \\ \hline
50 & 0.93000 & 0.93000 & 0.80000 & \textbf{0.990000} \\ \hline
100 & 0.820000 & 0.900000 & 0.860000 & 0.85000 \\ \hline
\end{tabular}
\label{tab:FMistral_Binary_results}
\end{table}
\subsubsection{Other LLMs for Attack Detection}
\begin{figure}[!ht]
    \centering
    \includegraphics[width=\linewidth]{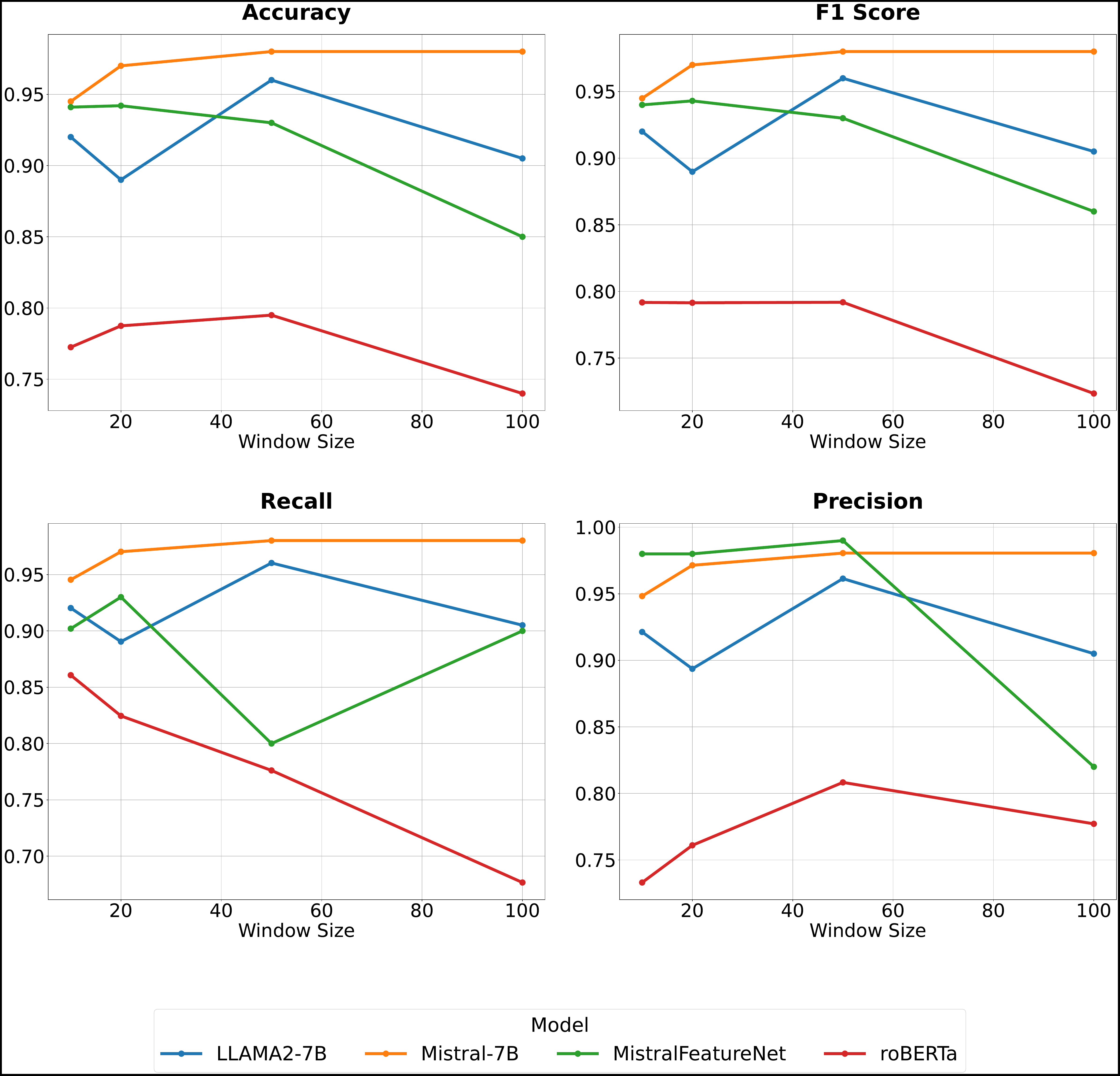}
    \captionsetup{justification=centering}
    \caption{Exploring Classification Metrics Variability among Binary Models}
    \label{fig: binary_classifers}
\end{figure}

To further validate the model choice, we fine-tuned RoBERTa and LLAMA2-7B under similar conditions. We employed similar qLoRA parameter described in Table\ref{tab:hyperparameters}. With LLAMA2-7B we obtained good results with only 1000 instance per class, whereas RoBERTa required a minimum of 3000 per class to converge.
Fig. \ref{fig: binary_classifers} summarizes the different models performance across the time windows. The immediate observation is the superiority of fine-tuned Mistral-7B over all the other models on most metrics, notably the accuracy and the F1 score. Conversely, RoBERTa, yields the worst results for all metrics. This is due to it being the only model with a predefined maximum sequence length of 512, i.e., RoBERTa cannot process token sequences of size surpassing 512. The most common solution to circumvent this limitation is to truncate the sequence to fit the model's requirements. However, most of our prompts surpass this value. The effect of the truncation is moderate for window size 10, however it becomes more prominent as the size of the prompts grows.  Another contributing factor to this behavior is the textualization step described in Section \ref{subsection: Veremi}, in which each column is considered as a sentence, so the truncation step results in eliminating whole features from the sequence.\\ 
Finally, Fig. \ref{fig: binary_classifers} suggests that 100 is the most suitable window size for the fine-tuned Mistral-7B. In contrast, 50 is the most suitable for LLAMA2-7B and MistralFeatureNet as they exhibit a deterioration when the window size is increased further.


\subsection{Misbehavior Classification}
 For misbehavior classification, the goal is to use Mistral-7B to detect and correctly classify attacks among the 9 classes described in Table \ref{tab:name_label_mapping}. We employ the identical reasoning as in the binary categorization. We fine-tune Mistral-7B, and then we compare its results against, MistralFeatureNet for the multi-class classification, LLAMA2-7B and RoBERTa.\\
\subsubsection{Fine-tuned Mistral-7B}
\begin{figure}[!ht]
    \centering
    \includegraphics[width=1\linewidth]{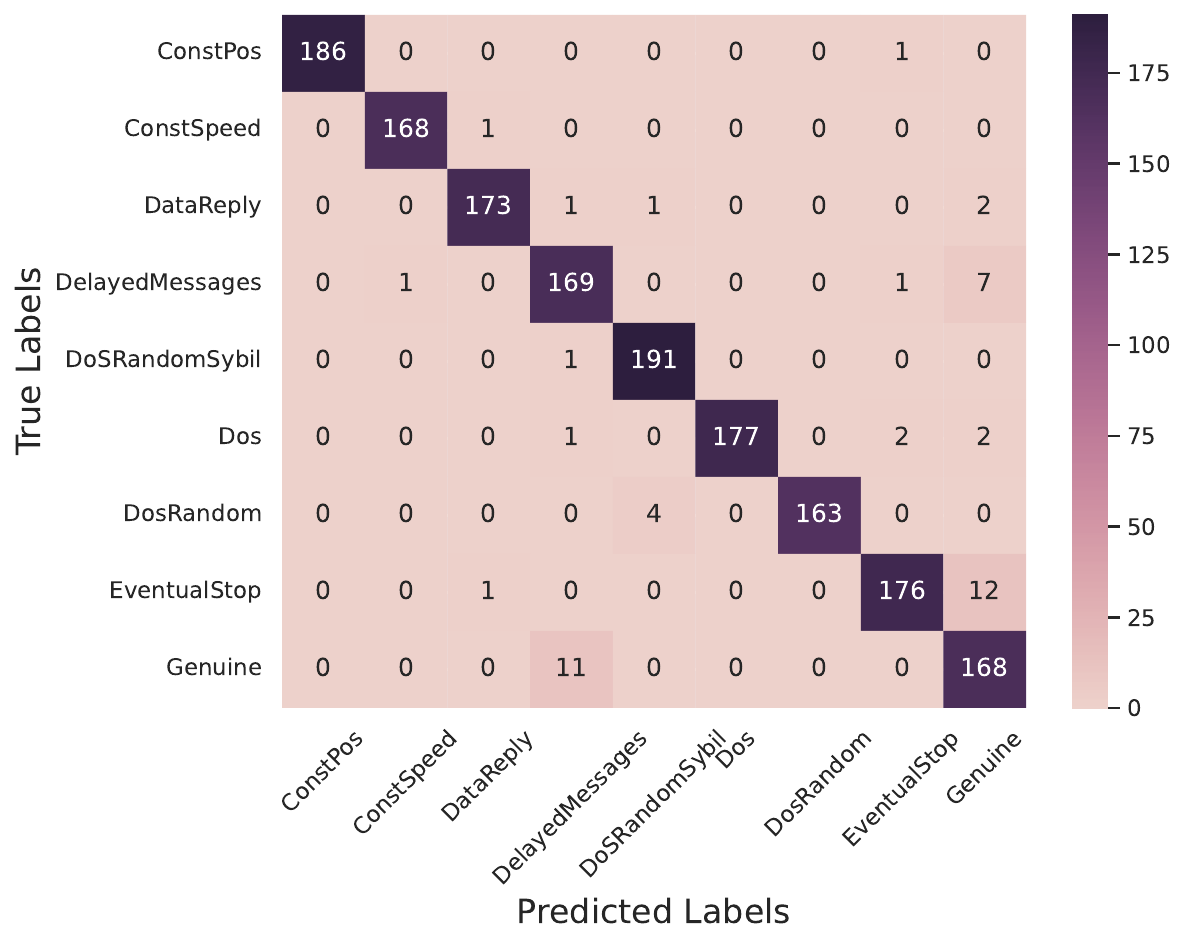}
    \captionsetup{justification=centering}
    \caption{Mistral-7B Confusion Metrics}
    \label{fig: Conf-Mtr}
\end{figure}

Similarly to the binary classification with Mistral-7B, we leverage qLoRA fine-tunning with similar hyperparameters, as they allow minimizing the computational cost. In this case, we select 1000 samples of each class to ensure the balance between the classes. \\ 
Fig. \ref{fig: multi_classifers} shows the performance results for different window sizes. As with binary classification, the hypothesis that the model performs better with a larger window is validated. The refined model obtained an F1 score of 0.96 in a 9-class classification task with just 1,000 instances per class. The confusion matrix in Fig. \ref{fig: Conf-Mtr} also proves the potential of the model, which correctly discerns most classes.\\

\subsubsection{Mistral For Feature Extraction}
Following the procedure of the binary classification, we freeze the layers of Mistral-7B, and we train a neuron network classifier leveraging the hyperparameters presented in Table \ref{tab:hyperparmeters_M_classifier}. Since the task involves a considerable number of classes, the convergence of the model required a bigger training set, thus we employed all the dataset. However, our data is unbalanced with a lot more benign instances, therefore, we undersampled the genuine class to  have a comparable representation to other classes.\\ 
The results, presented in Table \ref{tab:FMistral_Multiclass_results}, reflect the challenges encountered by the model in the 9-class classification task. The best outcome was scored with a window size of 20. However, the performance significantly drops for bigger window sizes. Therefore, for multiclass classification, fine-tuning is the best option to obtain good results. This is why we retained the fine-tuned. 

\begin{table}[!ht]
\centering
\caption{Classification network hyperparameters}
\begin{tabular}{|l|l|}
\hline
\textbf{Hyperparamter} & \textbf{Value} \\ \hline
Input dimension & 4096 \\ \hline
Hidden Layers dimension & {[}256, 128, 64, 32{]} \\ \hline
dropout rate & 0.2 \\ \hline
Optimizer & Adam \\ \hline
Learning rate & 0.0001 \\ \hline
Number of epochs & 150 \\ \hline
Batch size & 64 \\ \hline
\end{tabular}
\label{tab:hyperparmeters_M_classifier}
\end{table}

\begin{table}[!ht]
\centering
\caption{Multi-class MistralFeatureNet results}
\begin{tabular}{|c|c|c|c|c|}
\hline
Window & Accuracy & F1 Score & Recall & Precision \\ \hline
10 & 0.8692 & 0.8486 & 0.8692 & 0.8461 \\ \hline
20 & \textbf{0.8758} & \textbf{0.8537} & \textbf{0.8758} & \textbf{0.8490} \\ \hline
50 & 0.8385 & 0.8122 & 0.8385 & 0.8164 \\ \hline
100 & 0.6128 & 0.5654 & 0.6128 & 0.5587 \\ \hline
\end{tabular}
\label{tab:FMistral_Multiclass_results}
\end{table}

\subsubsection{Other LLMs for Attack Classification}
\begin{figure}[!ht]
    \centering
    \includegraphics[width=\linewidth]{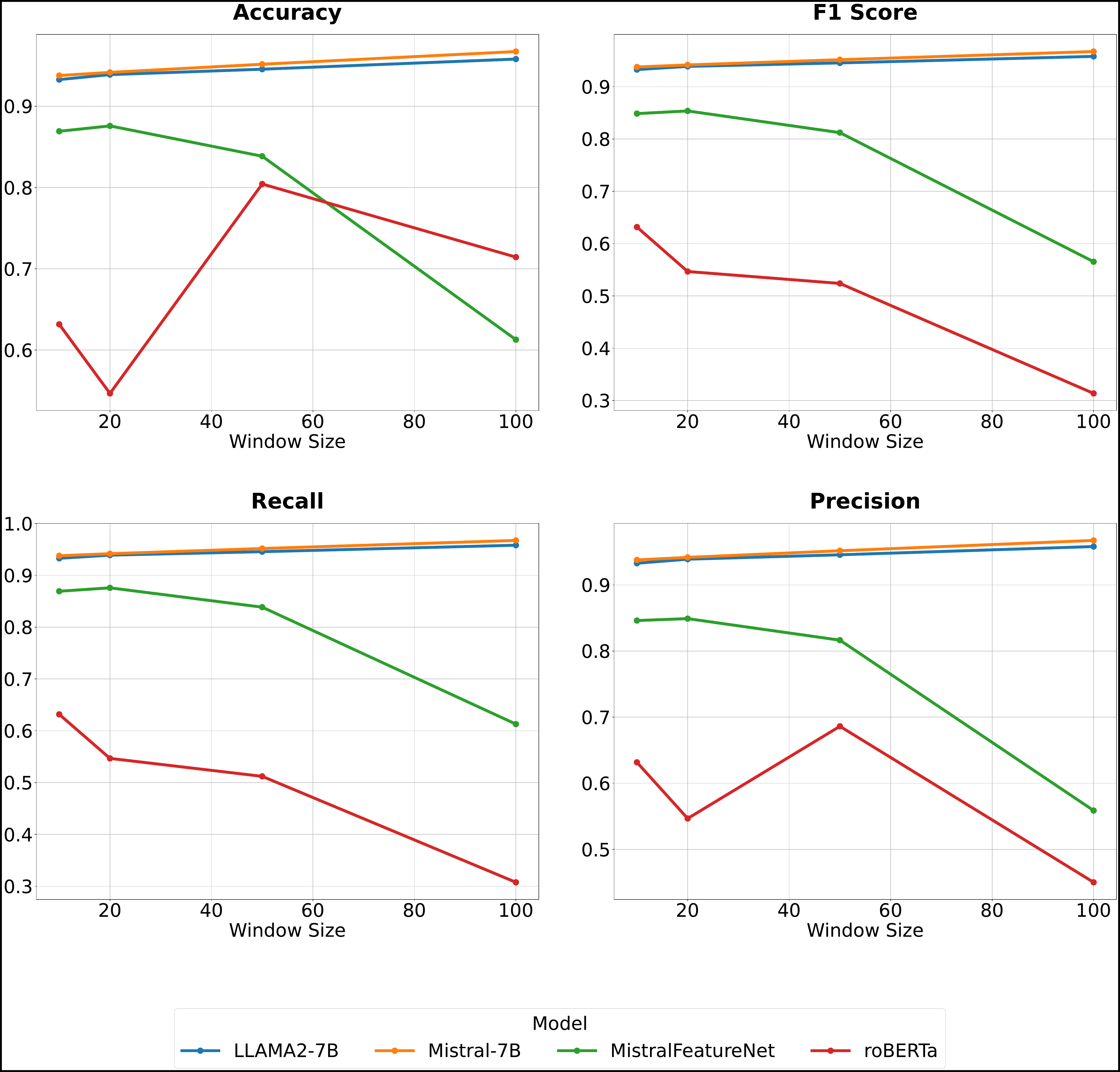}
    \captionsetup{justification=centering}
    \caption{Exploring Classification Metrics Variability among multiclass classifiers}
    \label{fig: multi_classifers}
\end{figure}

The performance of LLAMA2-7B and RoBERTa on the nine-class classification task is illustrated in Fig. \ref{fig: multi_classifers}. During training, LLAMA2-7B was configured similarly to Mistral-7B, with each class containing 1000 instances. In contrast, RoBERTa required a minimum of 4000 instances per class. The results highlight the superiority of large Transformers' over smaller ones. It is observed that LLAMA-7B achieves a similar accuracy to Mistral-7B, both around 0.95. Conversely, RoBERTa consistently exhibits the least beneficial outcomes. Finally, we find that larger models, particularly fine-tuned Mistral-7B, exhibit superior processing capabilities for complex data messages and consistently outperform smaller LLMs  such as RoBERTa in the multiple-class classification task. Moreover, they achieve this capacity with a reduced number of samples, a crucial advantage when  fewer vehicle communication data samples are available at the edge, which can occur due to fluctuations in traffic.

\subsection{Inference Time and Deployment Considerations}
To deploy MistralBSM, it is essential to inspect two key aspects: the computational resources required to store and run the model, and the inference latency. To address the first challenge, we propose deploying a fine-tuned and 4-bit quantized version of the model on the edge. This approach substantially reduces the memory footprint by 3$\times$, allowing the model to be stored within 4 GB of memory.

On the other hand, inference time is influenced by many factors, including the type of computational hardware, the size of the model, and the prompt length. In our case, the prompt is the BSM time window. Therefore, we focused our analysis on the impact of the window size on detection performance. 

\begin{figure}[!ht]
    \centering
    \includegraphics[width=0.8\linewidth]{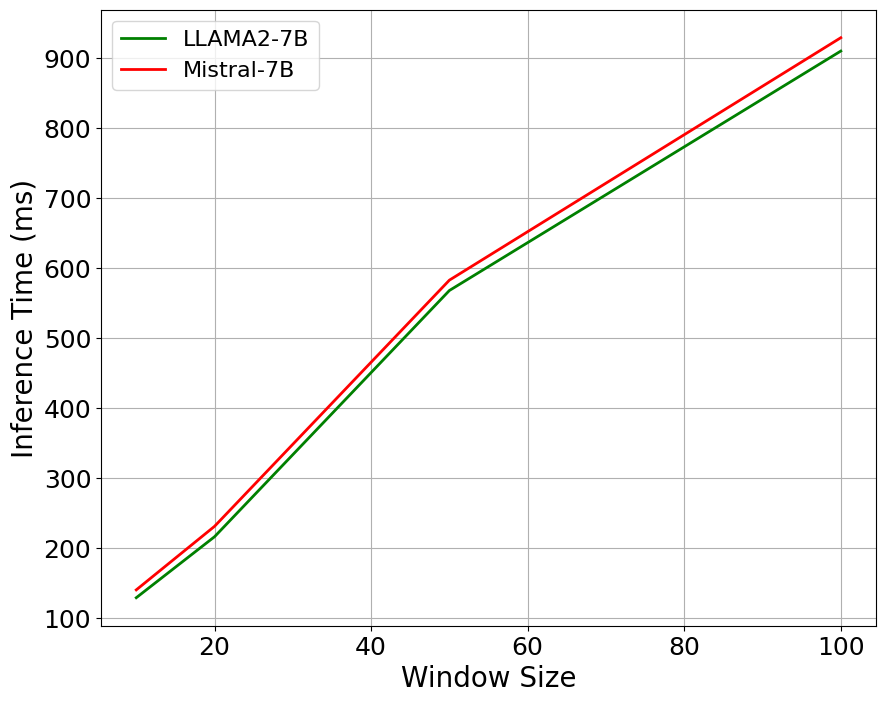}
    \captionsetup{justification=centering}
    \caption{Inference Time per Window Size}
    \label{fig:infer_time}
\end{figure}

The results discussed earlier show that for both binary and multiclass classification, larger window sizes generally yield better detection performance. With a 4$\%$ accuracy improvement when passing from 10 to 100 BSM per sequence. However, as illustrated in Figure \ref{fig:infer_time}, which depicts the inference time for different window sizes for Mistral-7B and LLaMA2-7B, inference time increases noticeably as window size increases achiving almost 1 second. This trend highlights a major constraint in applying LLMs for real-time misbehavior detection. To solve this issue, the edge-LLM, where latency is crucial, will process short time windows, while the cloud LLM will use longer sequences for better accuracy.

Despite the latency limitation, MistralBSM is designed to handle variable prompt lengths without requiring architectural modifications or additional preprocessing, ensuring flexibility in adapting the window size. As a consequence, the window size can be adjusted based on available computational resources to optimize performance without significant loss in accuracy.

\section{Conclusion}
\label{sec:conclusion}
In this paper, we explored the use of Large Language Models for vehicle misbehavior detection in vehicular networks, particularly by addressing the challenges of using LLMs at the edge, as part of a cloud-edge LLM based MDS system. We leveraged the new state-of-the-art LLM, Mistral-7B,  to detect abnormal behaving vehicles. Our detection process involves transforming a sequence of features extracted from BSMs into a text prompt, which is  then processed by a sequence classification mechanism using Mistral-7B. Through multiple experiments, we have demonstrated
the efficacy of fine-tuning Mistral-7B using QLoRA, even with
a limited dataset, across both binary and multiclass misbehavior classification  tasks. Furthermore, our comparative analysis revealed the
superior performance of Mistral-7B over concurrent LLMs,
including LLAMA2-7B and RoBERTa. 
Subsequently, we investigated the optimal number of examined packets 
or ”window size”, in a prompt to achieve the best results.
Our finding show that larger window sizes tends to enhance misbehavior classification.\\
In addition to evaluating model performance, we explored the practical considerations of deployment. We have addressed deployment challenges and proposed strategies such as quantization and the selection of an adequate window size to accommodate hardware constraints. 
Notably, our model's intrinsic flexibility with regard to varying window sizes, derived from its use of a language model, allows its adaptability to constraints of hardware configurations.
In conclusion, our study offers promising results as well as valuable practical insights for deploying misbehavior detection systems utilizing LLMs within vehicular networks contexts.


\end{document}